\documentclass{article}

\usepackage{microtype}
\usepackage{graphicx}
\usepackage{subcaption}
\usepackage{natbib}
\usepackage{booktabs} 

\usepackage{hyperref}

\usepackage{listings}
\usepackage{xcolor}
\usepackage{tikz}
\usetikzlibrary{shapes.geometric, arrows.meta,
positioning}

\usepackage[accepted]{icml2026}

\usepackage{amsmath}
\usepackage{amssymb}
\usepackage{mathtools}
\usepackage{amsthm}

\usepackage[capitalize,noabbrev]{cleveref}

\theoremstyle{plain}

\theoremstyle{definition}

\theoremstyle{remark}

\usepackage[textsize=tiny]{todonotes}

\icmltitlerunning{TreeAgent: A Generalizable Multi-Agent Framework for Automated Bias Labeling in Forestry}

\begin{document}

\twocolumn[
  \icmltitle{TreeAgent: A Generalizable Multi-Agent Framework for Automated Bias Labeling in Forestry via Compiled Expert Rules and Vision-Language Models}



  \icmlsetsymbol{equal}{*}

  \begin{icmlauthorlist}
    \icmlauthor{Shiyi Chen}{sch}
    \icmlauthor{Nicholas Saban}{sch}
    \icmlauthor{Collin Hargreaves}{sch}
    \icmlauthor{Huiqi Wang}{sch}
  \end{icmlauthorlist}

  \icmlaffiliation{sch}{University of California, Berkeley, California, United States}

  \icmlcorrespondingauthor{Huiqi Wang}{huiqiwang@berkeley.edu}

  \icmlkeywords{Machine Learning, ICML, AI4Science}

  \vskip 0.3in
]



\printAffiliationsAndNotice{}  

\begin{abstract}
Human-labeled data are widely used as reference annotations in ML, despite known variability across annotators in many expert-driven domains. In addition, expert annotation is slow, inconsistent, and remains a major bottleneck for scaling tasks like tree height bias classification in forestry remote sensing. We propose a multi-agent system (MAS) that orchestrates expert decision trees with Vision-Language Models (VLMs), treating the decision tree as a structural prior while VLMs perform localized semantic perception at individual nodes, with multi-agent voting to mitigate VLM stochasticity. We formalize a Decoupled Declarative Decision (D3) Framework that enables zero-modification generalization across diverse expert-defined decision structures. On a tree bias classification testbed, our framework outperforms supervised ML baselines and reduces the amount of expert labeling effort required. These results suggest that agentic orchestration of VLMs with expert priors can reproduce expert-defined labeling procedures at substantially lower annotation cost while maintaining interpretability.

\end{abstract}

\section{Introduction}
Tree height underpins terrestrial carbon accounting, biomass estimation, and 
the climate policies that depend on them~\citep{TOMPALSKI2014167, essd-17-965-2025}. 
Yet its three dominant data sources---field measurements, airborne lidar point 
clouds, and lidar-derived canopy height models (CHMs)---systematically 
disagree, each with distinct error modes: field crews struggle with tree tops 
in closed canopies, lidar misses apexes at low pulse density, and CHMs inherit both issues while adding gridding and interpolation artifacts. The true height of a tree is rarely directly observed, and the bias structure is heterogeneous across trees~\citep{WANG2019132, terrynGCB}.

To correct these biases at scale, domain experts hand-label individual trees using a rule-based diagnostic that distinguishes seven bias types from the geometry of measurement disagreement and local canopy context. Each tree takes 3--5 minutes of expert time, and inter-expert inconsistency injects noise into the very "reference" data that downstream ML correction models depend on. Across continental-scale inventories, however, this does not scale.

We argue this task is well suited to a multi-agent LLM system. The expert 
diagnostic is naturally decomposable: retrieve relevant measurements, reason 
over their numerical disagreement against a structured rule set, and inspect 
point-cloud or CHM figures only when the numbers are ambiguous. This 
decomposition lets us encode expert rules symbolically rather than hoping an 
LLM internalizes them from prompts, and invokes a vision-language model only 
at the perceptual steps where it is needed---yielding decisions auditable 
against the same rule book the experts use.

Naively encoding expert rules, however, creates a new bottleneck: symbolic 
classifiers tightly couple domain logic with implementation, so every 
threshold revision or taxonomy change requires re-engineering. We address 
this with the \textbf{Decoupled Declarative Decision (D3) framework}, which 
separates \emph{what} to decide from \emph{how} to decide it. D3 has two 
components: a \emph{Logic Primitive Inventory} (LPI) defining a closed 
vocabulary of atomic execution primitives, and a \emph{Neural Rule Transpiler} 
(NRT) that compiles natural-language expert rules into executable, 
validatable decision graphs over the LPI~\citep{lpi, gorski2025integratingexpertknowledgelogical}. 
Rule revisions become configuration edits; every decision is traceable to 
the rule that produced it.

Concretely, we ask: \textbf{(RQ1)} Can a VLM reliably detect the perceptual 
cues, canopy overlap, ground outliers, that experts rely on? 
\textbf{(RQ2)} Can a multi-agent system, given expert rules, classify the 
seven bias types end-to-end? \textbf{(RQ3)} Does the resulting system outperform conventional ML 
classifiers on this task?

We instantiate D3 as \textbf{TreeAgent}, a multi-agent system that answers 
these questions affirmatively. Our contributions are: \textbf{(C1)} the 
\emph{D3 framework} for decoupling expert classification logic from 
execution, with a closed LPI and an LLM-based NRT that compiles expert 
rules into validated, auditable decision graphs; \textbf{(C2)} \emph{TreeAgent}, 
an instantiation encoding a forestry expert's seven-class bias diagnostic, 
with a majority-vote VLM agent handling the graph's perceptual nodes; and 
\textbf{(C3)} an \emph{empirical evaluation} on expert-labeled trees from 
two held-out NEON sites showing TreeAgent reaches 67.6\% Macro-F1 at 
$\sim$0.040 minutes per tree, against a tuned tabular ML baseline at 36.2\% 
macro-F1 despite extensive feature engineering, imbalance correction, and 
tabular foundation-model substitution.

Beyond forestry, D3 offers a general recipe for scientific labeling 
workflows where expert reasoning is structured, evolving, and requires 
occasional perceptual judgment---a regime where end-to-end models discard 
interpretability and single-prompt LLMs discard reliability.

\section{Related Work}
\label{sec:related}

\paragraph{LLMs as compilers for expert knowledge.} A growing line of work
integrates expert knowledge into LLM-driven pipelines as structured programs
rather than free-form prompts \citep{gorski2025integratingexpertknowledgelogical}.
D3 sits in this lineage but enforces a closed primitive inventory at compile
time, so every executable graph is statically validatable before deployment.

\paragraph{Tabular ML and small-data scientific labeling.} Gradient-boosted
trees remain the dominant model family on small tabular tasks
\citep{breiman2001random,chen2016xgboost,ke2017lightgbm,grinsztajn2022tree};
recent tabular foundation models extend this regime with in-context learning
\citep{hollmann2025tabpfn}. We use these as the supervised baseline and show
that even with imbalance correction \citep{chawla2002smote, lin2017focal} and
transductive adaptation \citep{lee2013pseudo}, a tuned ensemble does not match
the agent system on the seven-class target.

\paragraph{Forestry remote sensing.} The NEON Airborne Observation Platform and
the broader literature on LiDAR-derived canopy products document the standard
data products and their uncertainties
\citep{duncanson2015importance,thorpe2016introduction}; per-tree bias-source
classification has not previously been attempted at scale.

\section{Methods}
\subsection{The D3 Framework}
\label{sec:d3}

\begin{figure*}
  \vskip 0.1in
  \begin{center}
    \centerline{\includegraphics[width=\textwidth,trim=0 10 0 80,clip]{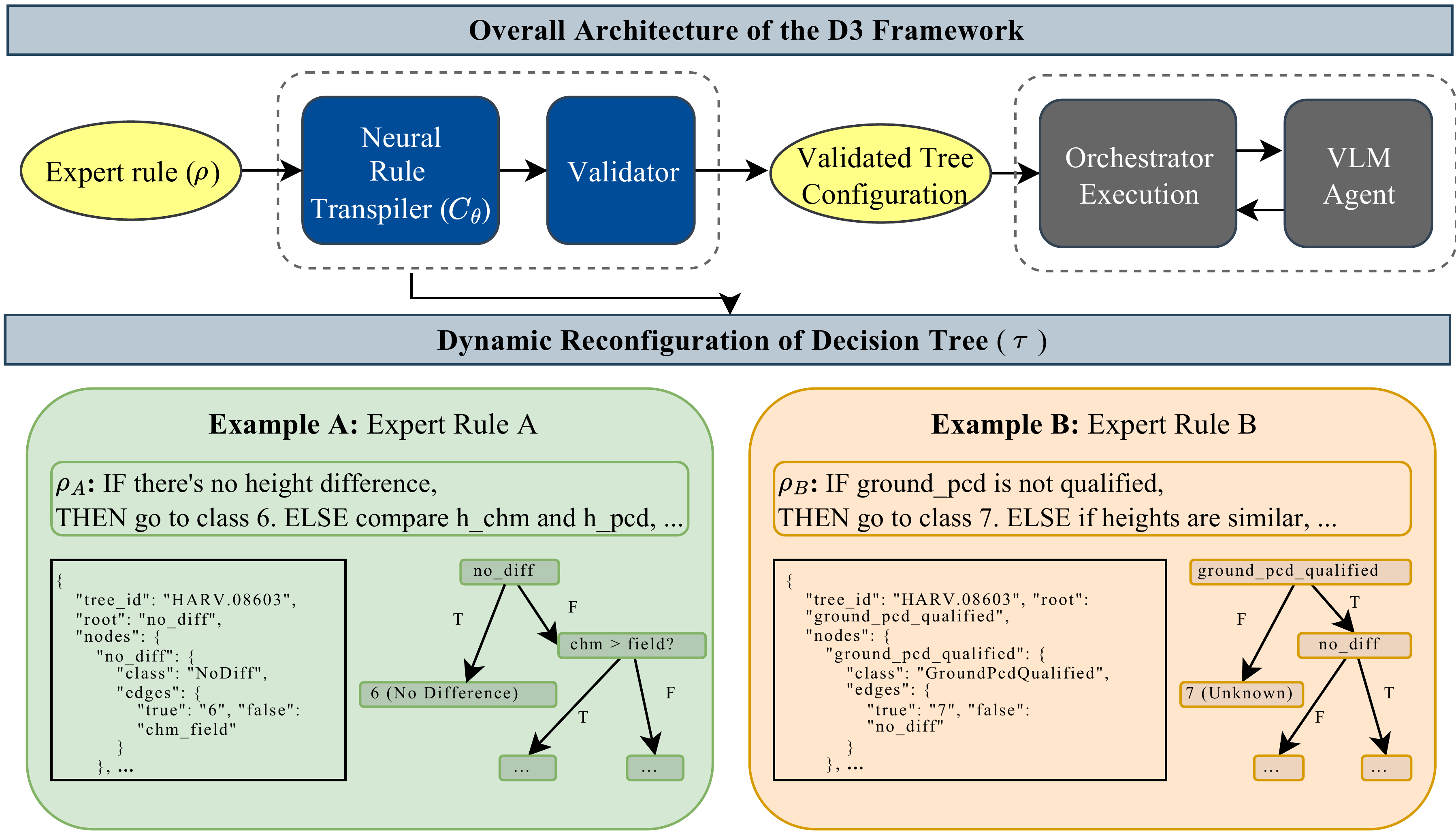}}
    \caption{
      \textbf{Overview of the D3 Framework.} The architecture decouples
domain-specific logic through a fixed Logic Primitive Inventory (LPI).
\textbf{Top:} the Neural Rule Transpiler (NRT) translates an unstructured
expert rule $\rho$ into a structured JSON tree configuration $\mathcal{T}$ via
a single inference call. \textbf{Bottom:} distinct expert strategies
($\rho_A, \rho_B$) are compiled into different executable tree structures
without altering the underlying Orchestrator or VLM Agent code, achieving
zero-modification generalizability.}
    \label{fig:d3}
  \end{center}
\end{figure*}

Continuous efforts have been made to integrate expert knowledge into structured forms using LLMs~\citep{gorski2025integratingexpertknowledgelogical}. We propose \textbf{Decoupled Declarative Decision (D3) Framework}, which, as the architectural backbone of TreeAgent, adopts and expands this idea to also include VLMs. D3 separates \emph{what} a domain expert wants to decide from \emph{how} the system executes that decision. The expert writes a rule $\rho$ in structured
natural language; D3 compiles it into a typed graph $\mathcal{T}$ over a fixed, finite \textbf{Logic Primitive Inventory} ($\mathcal{V}$); and an orchestrator runs $\mathcal{T}$ on incoming samples (Figure~\ref{fig:d3}). Two practical properties follow: \textbf{verifiability} --- every classification is traceable to the rule node that produced it, since the rule is compiled into an inspectable graph rather than absorbed into model weights --- and \textbf{zero-modification generalizability} --- when an expert revises a condition or the diagnostic flow, the change is a configuration edit, not a code change.

\subsubsection{Logic Primitive Inventory (LPI)}
\label{sec:lpi}

The LPI defines a fixed, finite set of node classes
$\mathcal{V} = \mathcal{V}_{\text{det}} \cup \mathcal{V}_{\text{vlm}} \cup \mathcal{V}_{\text{exit}}$,
partitioned by \emph{execution type} $\omega(v) \in \{\texttt{det},\,\texttt{vlm},\,\texttt{exit}\}$.
The execution type determines \emph{how} the orchestrator will evaluate a node:
\texttt{det} nodes are resolved by a closed-form arithmetic predicate;
\texttt{vlm} nodes are resolved by querying the vision-language model agent;
\texttt{exit} nodes terminate execution and return a class label.
Each node class $v \in \mathcal{V}$ is a tuple

\begin{equation}
    v = \bigl\langle\,\mathrm{id}(v),\;\omega(v),\;\varepsilon(v),\;\vartheta(v)\bigr\rangle,
    \label{eq:node-tuple}
\end{equation}

where $\varepsilon(v) \subseteq \mathcal{F} \cup \mathcal{I}$ is the \emph{expected input signature}---the subset of tabular metric fields $\mathcal{F}$ and/or image modalities $\mathcal{I}$ that the node's evaluation kernel requires---and
$\vartheta(v)$ is the \emph{evaluation kernel}: a predicate or prompt whose
output is always binary ($\{0,1\}$), forming the branching condition
that routes the traversal to one of two successor nodes.

\textbf{Deterministic nodes} ($\mathcal{V}_{\text{det}}$) evaluate a closed-form predicate over $\varepsilon(v) \subseteq \mathcal{F}$, giving an exact,
side-effect-free outcome $o_v = \vartheta(v)(x_{\varepsilon(v)}) \in \{0,1\}$.
\textbf{VLM nodes} ($\mathcal{V}_{\text{vlm}}$) dispatch $\vartheta(v)$ as a
natural-language prompt to a vision-language agent operating on the image
modalities in $\varepsilon(v)$; the agent returns a binary outcome.
\textbf{Exit nodes} ($\mathcal{V}_{\text{exit}}$) carry a class label
$\vartheta(v) \in \mathcal{L}$ and terminate execution.

\subsubsection{Tree Configuration}
\label{sec:tree-config}

A \emph{tree configuration} $\mathcal{T}$ is a directed acyclic graph instantiated
over $\mathcal{V}$:
\begin{equation}
    \mathcal{T} = \langle\,\texttt{tree\_id},\;r,\;N,\;E\,\rangle,
    \label{eq:tree-config}
\end{equation}
where $\texttt{tree\_id}$ uniquely identifies the configuration,
$r \in N$ is the root node, $N$ is a multiset of node instances drawn from
$\mathcal{V}$, and
\begin{equation}
    E \subseteq N_{\lnot\text{exit}} \times \{0,1\} \times N
    \label{eq:edge-set}
\end{equation}
is the edge relation mapping each non-exit node and binary outcome to a unique
successor. A valid $\mathcal{T}$ satisfies two well-formedness conditions:
(1)~\textbf{Completeness}: every $n \in N_{\lnot\text{exit}}$ has exactly one
outgoing edge for each outcome in $\{0,1\}$; and
(2)~\textbf{Vocabulary closure}: the class of every $n \in N$ is an element of
$\mathcal{V}$, enforced by a deterministic post-compilation validator
(Section~\ref{sec:nrt}).

\subsubsection{Neural Rule Transpiler (NRT)}
\label{sec:nrt}

Given an expert rule $\rho$ expressed in structured natural language and a tree
identifier \texttt{tree\_id}, the NRT produces a tree configuration:
\begin{equation}
    \mathcal{C}_{\theta} : (\rho,\;\texttt{tree\_id}) \;\longmapsto\; \mathcal{T},
    \label{eq:nrt}
\end{equation}
where $\mathcal{C}_{\theta}$ is an LLM with parameters $\theta$, prompted with the full vocabulary $\mathcal{V}$ and the JSON schema of $\mathcal{T}$.
The model is constrained to output only the JSON serialisation of $\mathcal{T}$; the
output space is therefore restricted to $\{\mathcal{T} : \forall n \in N,\, \mathrm{class}(n) \in \mathcal{V}\}$. Compilation is a single inference call with no
agentic loop. Appendix~\ref{app:d3} includes the prompts for the NRT. 

Two properties follow from this design.
\textbf{Verifiability.} Because $\mathcal{V}$ is finite and known at compile time, the validator runs in $O(|N|)$ by checking class membership for each node. Any
$\mathcal{T}$ that fails validation is rejected before reaching the orchestrator, guaranteeing that only structurally valid trees are executed.
\textbf{Semantic alignment without schema exposure.} $\mathcal{C}_{\theta}$ performs
implicit field grounding: natural-language variable references in $\rho$ (e.g.\
``field-measured tree height'') are mapped to canonical identifiers in $\mathcal{F}$
(e.g.\ \texttt{h\_field}) as a by-product of compilation. The interface to the domain expert is therefore unrestricted prose $\rho$; the interface to the orchestrator is a fully typed, validated $\mathcal{T}$. No schema knowledge is required of the expert.

\subsubsection{Orchestrator Execution}
\label{sec:orchestrator}
The orchestrator traverses $\mathcal{T}$ given a sample
$\mathbf{x} = (\mathbf{x}_{\mathcal{F}},\,\mathbf{x}_{\mathcal{I}})$:
\begin{equation}
    n_{t+1} = \mathrm{succ}(n_t,\; o_{n_t}(\mathbf{x})),
    \label{eq:traversal}
\end{equation}

\begin{equation}
    o_{n_t}(\mathbf{x}) =
    \begin{cases}
        \begin{aligned}
            &\vartheta(n_t)\!\left(\mathbf{x}_{\varepsilon(n_t)}\right) \\
            &\quad\text{if } \omega(n_t) = \texttt{det}, \\[4pt]
            &\mathcal{A}_{\mathrm{vlm}}\!\left(\vartheta(n_t),\,
                \mathbf{x}_{\varepsilon(n_t)},\, \texttt{tree\_id}\right) \\
            &\quad\text{if } \omega(n_t) = \texttt{vlm},
        \end{aligned}
    \end{cases}
    \label{eq:traversal1}
\end{equation}
where $\mathrm{succ}(n,o)$ looks up the successor of node $n$ under outcome $o$
in $E$. Execution terminates when $\omega(n_t) = \texttt{exit}$, at which point
$\vartheta(n_t) \in \mathcal{L}$ is returned as the classification label. The
orchestrator contains no expert rules and is entirely agnostic to the structure of
$\mathcal{T}$.

\subsubsection{Configuration Generalizability}
\label{sec:generalizability}

The central property of the D3 Framework is zero-modification reconfigurability within a domain whose decision logic can be expressed using the available primitive inventory:
for any expert rule $\rho'$ from the same domain, the full pipeline
\begin{equation}
    \rho' \;\xrightarrow{\mathcal{C}_\theta}\; \mathcal{T}' \;\xrightarrow{\text{orchestrator}}\; \hat{y}
    \label{eq:generalizability}
\end{equation}
requires no changes to $\mathcal{V}$, $\mathcal{C}_\theta$, the validator, or the
orchestrator, provided $\rho'$ can be expressed as a binary decision process over
$\mathcal{F} \cup \mathcal{I}$. This holds because the orchestrator's traversal
(Eq.~\ref{eq:traversal1}) is parameterised entirely by $\mathcal{T}$, not
hard-coded to any particular rule.

To evaluate zero-modification reconfigurability independently of end-to-end
performance, we authored five expert rules in natural language with varying root nodes, branching order, and subsets of
$\mathcal{V}$ classes, and passed
each to the NRT in a single inference call. All compilations produced tree configurations that passed both manual inspection and the automated vocabulary-closure / completeness validator described in Section~\ref{sec:nrt}.

\subsection{The VLM Agent Architecture}
\label{sec:vlm}

\begin{figure}
  \vskip 0.1in
  \begin{center}
    \centerline{\includegraphics[width=\columnwidth]{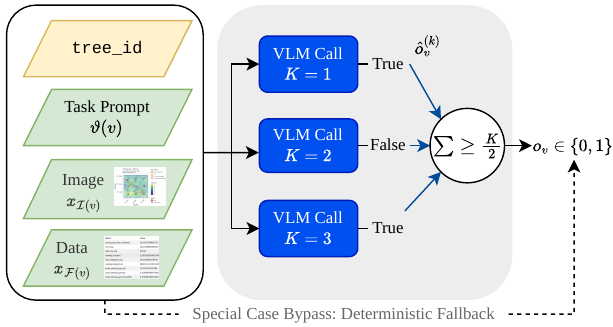}}
    \caption{
      \textbf{The VLM agent.} Each VLM node receives a node-specific prompt
$\vartheta(v)$, image modalities, and structured fields. To suppress
stochasticity, $K{=}3$ independent samples at temperature $\tau{=}0.2$ are
aggregated by majority vote. Two node classes use deterministic fallbacks when
the input alone is sufficient.
    }
    \label{fig:vlm}
  \end{center}
\end{figure}
\paragraph{VLM agent.} Each VLM node delegates to a vision-language agent
$\mathcal{A}_\text{vlm}$ that receives the prompt $\vartheta(v)$, image inputs
$x_{\mathcal{I}(v)}$, and the \texttt{tree\_id}. To mitigate sampling variance,
we wrap the base model in a $K=3$ majority vote at temperature $\tau=0.2$:
$o_v = \mathbf{1}\bigl[\sum_k \hat o_v^{(k)} \geq K/2\bigr]$, with each
$\hat o_v^{(k)} \in \{0,1\}$ parsed by regex from the model's response
(Figure~\ref{fig:vlm}).

\paragraph{VLM Input Figures.} For each tree the agent receives (i) a bird's-
eye CHM window with crown polygons drawn as dashed circles, stems as brown
dots, and canopy height as a color gradient; and (ii) a two-panel transect
cross-section through the stem in the N--S and E--W directions, with returns
colored by LiDAR class.





\subsection{VLM Prompt Design}
\label{sec:vlm-prompt}
Two VLM node classes require visual judgment that cannot be reduced to a formula. Appendix~\ref{app:prompts} includes the prompts for the two VLM nodes. 

\texttt{CrownOverlap} requires deciding whether a neighboring tree's dashed crown-delineation circle intersects the target tree's crown in a bird's-eye CHM rendering, a judgment that depends on visual context (relative crown sizes, partial occlusion, plot edge effects) and cannot be reduced to a closed-form predicate over the available tabular fields.

\texttt{GroundPcdOutlier} checks whether the ground elevation for a tree in a given year is abnormally high compared to other survey years. A false high ground reading compresses the tree height and requires inspection of a time series transect image of ground elevation across years to spot the outlier.

We test four prompt versions on \texttt{CrownOverlap} to determine how much guidance the model needs to make this call reliably.

\paragraph{VLM Input Figures.} Each tree in the evaluation set is represented by
two figures passed as visual context:

\textbf{(i)}~a bird's-eye canopy height model (CHM)
window, with individual tree crown boundaries shown as dashed circles, stem
locations as brown dots, and canopy height as a color gradient.

\textbf{(ii)}~a two-panel transect cross-section showing the full point
cloud in the N--S and E--W directions through the target stem, with returns
colored by LiDAR class (ground, low vegetation, high vegetation) and the
field-survey measurement point annotated as a red circle.

\paragraph{VLM Prompt variants.}
We test four prompt versions of increasing detail; full text is in
Appendix~\ref{app:prompts}.
Table~\ref{tab:prompt-variants} shows which components each includes.

\textbf{Minimal} is a bare question with no system context.

\textbf{V1} adds a system prompt explaining the two figures (color scale, dashed
crown circles, stem dots).

\textbf{V2} keeps the V1 system prompt and adds a five-step chain-of-thought
\citep[p.~2]{wei2022chain}: read axes, locate the target crown, check for circle intersection in
Figure~1, compare neighbor heights, and inspect the Figure~2 transect.

\textbf{V3} replaces the system prompt with expert framing
(``experienced forestry and LiDAR analyst'') and provides the most detailed
per-figure reading instructions, with explicit disambiguation of crown-circle
overlap versus stem-circle containment.

\paragraph{Test setup.}
We test all four prompt versions on 50 OSBS trees sampled to cover 34
plot--year combinations (random seed 42), ensuring the test set spans different
plots and survey years rather than drawing from a single context.
Labels were assigned manually and marked for crown overlap, yielding 22
positives and 28 negatives.
We report overall accuracy, recall (true-positive rate), and specificity
(true-negative rate).

\begin{table}[!h]
  \caption{Components present in each \texttt{CrownOverlap} prompt variant.
           $\checkmark$ = included; $-$ = absent.}
  \label{tab:prompt-variants}
  \begin{center}
    \begin{small}
      \begin{tabular}{lcccc}
        \toprule
        & Figure & Chain-of- & Expert & \texttt{taller} \\
        Variant & description & thought & framing & in output \\
        \midrule
        Minimal & $-$          & $-$          & $-$          & $-$          \\
        V1      & $\checkmark$ & $-$          & $-$          & $-$          \\
        V2      & $\checkmark$ & $\checkmark$ & $-$          & $\checkmark$ \\
        V3      & $\checkmark$ & $\checkmark$ & $\checkmark$ & $\checkmark$ \\
        \bottomrule
      \end{tabular}
    \end{small}
  \end{center}
\end{table}

\begin{table}[h]
    \caption{Step-by-step reasoning instructions given to the model in V2 and V3.}
    \label{tab:cot-steps}
    \begin{center}
      \begin{small}
        \resizebox{\columnwidth}{!}{%
        \begin{tabular}{cl}
          \toprule
          Step & Action \\
          \midrule
          1 & Read Figure 1 axes and canopy height color scale \\
          2 & Locate the target tree's crown circle in the bird's-eye map \\
          3 & Check whether any neighboring crown circle overlaps the target \\
          4 & If overlap: is that neighbor's canopy visually taller? \\
          5 & Check Figure 2 cross-section for canopy returns above the target apex \\
          \bottomrule
        \end{tabular}%
        }
      \end{small}
    \end{center}
  \end{table}

\subsection{Supervised ML Baseline}
\label{sec:ml-baseline}

We train a tabular classifier on hand-engineered features summarizing each tree's three height sources and its local point-cloud and DTM context, and use it as the supervised baseline against which we compare TreeAgent. The baseline answers a question distinct from the agent's: \emph{how far does a small-data expert-feature pipeline go on its own?} Its per-class errors also identify which decisions in the expert diagnostic resolve from summary statistics and which require the spatial perception that VLM nodes provide.

\paragraph{Bias taxonomy.} Each tree carries one of seven labels assigned by an expert from the disagreement pattern among three height sources: $H_{\mathrm{field}}$ (the field-based height), $H_{\mathrm{chm}}$ (the canopy height model raster value), and $H_{\mathrm{pcd}}$ (the maximum point-cloud return inside the crown buffer). All these three values are absolute elevation, with base of tree elevation added. The labels are: \texttt{bias\_1}, $H_{\mathrm{field}}$ underestimation (the survey reads short); \texttt{bias\_2}, $H_{\mathrm{field}}$ overestimation (the survey reads tall); \texttt{bias\_3}, $H_{\mathrm{chm}}$ overestimation (the CHM is inflated by a taller neighboring canopy); \texttt{bias\_4}, $H_{\mathrm{chm}}$ underestimation with $H_{\mathrm{pcd}}$ matching the field height (the rasterized CHM misses the apex but the raw LiDAR points capture it); \texttt{bias\_5}, both $H_{\mathrm{chm}}$ and $H_{\mathrm{pcd}}$ underestimate; \texttt{no\_bias}, all three sources agree; and \texttt{unknown}, the expert could not place the tree in any of the above categories within their five-minute labeling budget.

\paragraph{Data and split.} The dataset contains 283 expert-labeled trees from three NEON sites~\citep{thorpe2016introduction}: OSBS (Florida pine-oak), WREF (Washington temperate conifers), and SRER (Arizona desert shrubs). We train on all 136 OSBS trees and test on the 147 WREF and SRER trees, predicting the full seven-class taxonomy. The split is geographic by design: it tests whether a classifier trained in one ecosystem generalises to structurally dissimilar ones, which is the operationally relevant question for continental-scale deployment. We retain the \texttt{unknown} class as a target so the classifier predicts over the same label space the experts produced.

\paragraph{Features.} We summarize each tree with 25 features grouped by source. From the NEON Vegetation Structure field survey~\citep{thorpe2016introduction}: four numeric measurements (\texttt{height}, \texttt{stemDiameter}, \texttt{crownRadius}, \texttt{adjElevation}) and four categorical descriptors (\texttt{canopyPosition}, \texttt{plantStatus}, \texttt{growthForm}, \texttt{taxonID}). From the canopy height model raster: \texttt{chm\_height}, the single-pixel CHM value at the field-survey coordinate, and \texttt{chm\_height\_buff}, the maximum CHM value within a circular buffer of radius $0.9 \times \texttt{crownRadius}$ (minimum 1\,m) centered on the stem; the buffered version recovers the canopy apex when the stem is offset from the tallest pixel, as is common for leaning trees. From the LiDAR point cloud: \texttt{Z\_pointcloud} (the mean elevation of returns classified as ground, i.e.\ LAS classification code 2, inside the per-tree buffer; this is a local terrain height); \texttt{std\_pointcloud} (the standard deviation of those ground-point elevations, a local terrain roughness); \texttt{pointdensity\_ground\_tree} (the count of ground points per unit area); and three Digital Terrain Model (DTM) quality measures: \texttt{dtm\_nn} (the elevation reported by the nearest defined DTM grid cell), \texttt{dtm\_nn\_dist} (the distance to that cell, where large values flag unreliable terrain interpolation), and \texttt{dtm\_buffer} (the mean DTM elevation inside the per-tree buffer). The remaining nine features are derived ratios and differences (Appendix~\ref{app:features}): notably \texttt{diff\_chm\_survey} ($H_{\mathrm{chm}} - H_{\mathrm{field}}$), \texttt{diff\_chm\_buff\_survey}, \texttt{ground\_elevation\_diff}, and \texttt{height\_to\_crown\_ratio}.

\paragraph{Models.} LightGBM~\citep{ke2017lightgbm} is the primary model, motivated by the dominance of gradient-boosted decision trees on small tabular tasks~\citep{grinsztajn2022tree}. Class weights are set to inverse class frequency to handle the 4-to-46 sample imbalance across the seven classes. Hyperparameters are fixed (maximum depth 3, 300 boosting rounds, $L_1/L_2$ regularization $=2.0$, subsample $=0.7$); we do not cross-validate given $n_{\text{train}}=136$. Appendix~\ref{app:ml-ablations} reports XGBoost~\citep{chen2016xgboost}, Random Forest~\citep{breiman2001random}, CatBoost, HistGradientBoosting, a stacking ensemble (LightGBM, RF, and XGBoost feeding a logistic regression meta-learner with out-of-fold predictions), and the TabPFN tabular foundation model~\citep{hollmann2025tabpfn} as model-family ablations.

\label{sec:pointnet}
\paragraph{Point cloud as features.} The hand-engineered features encode substantial expert knowledge: which CHM neighbourhood to read (buffered max, not point), which ground-quality summaries discriminate, and how to enforce scale-invariance across ecosystems. We isolate the value of this expert pipeline by asking the complementary question: can a learned representation, trained end-to-end on the same raw point clouds, recover this performance from labels alone? A negative result supports the framing of D3, which assumes the expert rules carry information small-data learning cannot easily extract. For each tree we crop the LAS tile to a circular buffer at the field-survey coordinate  (radius $= \max(\texttt{crownRadius}, 2\text{m})$), then sample or pad to 1024 points, with the LAS classification code retained as additional feature alongside. Extraction succeeds for all 283 trees. The model is a stripped-down PointNet~\citep{qi2017pointnet} with three Conv1D layers, batch normalisation, max-pool over points, two FC layers, and dropout; we omit the T-Net because it overfits at this scale. Training uses AdamW with cosine LR decay, balanced class weights, and augmentation by random rotation around the vertical axis, jitter, and 5\% point dropout. We grid-search hidden width $\in \{32, 64, 128\}$ and learning rate $\in \{10^{-3}, 3\times 10^{-4}\}$ for 80 epochs each. The split and target match the supervised baseline.

\section{Results}
\subsection{TreeAgent Evaluation}
\label{sec:treeagent-eval}

We evaluate TreeAgent on 147 expert-labeled trees from the SRER and WREF
sites drawn from the NEON API, treating human-assigned bias labels as reference labels produced under the expert diagnostic protocol. We compare against the supervised ML baseline of
Section~\ref{sec:ml-baseline} and the single-VLM prompt results of
Section~\ref{sec:vlm-eval}.

TreeAgent's behaviour depends on the natural-language rule passed to the
NRT, and a single expert diagnostic admits several semantically plausible
phrasings of the \texttt{NoDiff} tolerance --- $<$ vs $\le$, and
normalisation by $H_{\mathrm{field}}$ (elevation of the tree apex) vs \texttt{survey\_height} (tree height value). To avoid
conditioning the headline on a post-hoc rule choice, we freeze a single
rule \emph{a priori} on procedural grounds and report that as the
headline, then report the three remaining variants as a sensitivity check.
The frozen-rule layer is the apples-to-apples comparison against the
LightGBM baseline.

\paragraph{Headline rule (chosen \emph{a priori}).}
The human annotators used a tolerance of 2\% of the survey height
with non-strict inequality when assigning the \texttt{NoDiff} label. We
therefore freeze
\[
  (\{H_{\mathrm{chm}}, H_{\mathrm{pcd}}\} - H_{\mathrm{field}}) /
  \texttt{survey\_height} \;\le\; 0.02
\]
as the headline rule because it matches the annotation procedure that
generated the ground-truth labels. We emphasise that this rule was
selected for procedural fidelity, \emph{not} because it maximised any
test metric.

\paragraph{Headline result.}
Under the frozen rule, TreeAgent reaches
\textbf{67.6\%} Macro-F1 with an average per-tree runtime
of \textbf{0.040} minutes (Table~\ref{mas-acc}), against
36.2\% Macro-F1 for the tuned LightGBM baseline
(Section~\ref{sec:ml-headline}) and roughly 5 minutes per tree for human
annotation. Per-class recalls under this rule are in
Table~\ref{per-class-acc}.

\begin{table}[!h]
  \caption{TreeAgent under the frozen headline rule, against the supervised baseline and human annotation (reference). Human labeling is included as a time-cost reference baseline only. }
  \label{mas-acc}
  \begin{center}
    \begin{small}
      \begin{sc}
      \resizebox{\columnwidth}{!}{%
        \begin{tabular}{lcc}
          \toprule
          Method                              & Time (min)              & Macro-F1 \\
          \midrule
          TreeAgent (frozen rule)             & 0.040\,$\pm$\,0.007 & 67.6\% \\
          Supervised ML (LightGBM) & N/A                     & 36.2\% \\
          Human Labeling (reference) & 5.0\,$\pm$\,2.0         & N/A \\
          \bottomrule
        \end{tabular}%
        }
      \end{sc}
    \end{small}
  \end{center}
\end{table}

\begin{table}[!h]
  \caption{Per-class recall under the frozen headline rule. \texttt{n} is
  the number of test trees in each class.}
  \label{per-class-acc}
  \begin{center}
    \begin{small}
      \begin{sc}
      \resizebox{\columnwidth}{!}{%
        \begin{tabular}{lcc}
          \toprule
          Bias class & Recall & n \\
          \midrule
          1 \texttt{H\_field} underestimation              & 72.2$\pm{17.2}$\% & 18 \\
          2 \texttt{H\_field} overestimation               & 65.9$\pm{9.0}$\% & 41 \\
          3 \texttt{H\_chm} overestimation                 & 71.4$\pm{25.2}$\% & 7  \\
          4 \texttt{H\_chm} underestimation only           & 66.7$\pm{20.7}$\% & 9  \\
          5 \texttt{H\_chm} and \texttt{H\_pcd} underestimation & 66.7$\pm{16.0}$\% & 12 \\
          6 no difference                                  & 85.7$\pm{0.0}$\% & 14 \\
          7 unknown                                        & 72.2$\pm{8.6}$\% & 46 \\
          \bottomrule
        \end{tabular}%
        }
      \end{sc}
    \end{small}
  \end{center}
\end{table}

\paragraph{Human-label variability.} We acknowledge that in practice, trees in complex situations may receive different labels from experts with varying domain experience and training quality. The labels are therefore better viewed as expert reference annotations than as objective ground truth. Since our evaluation assumes a single expert annotation per tree, the reported Macro-F1 should be interpreted as agreement with one expert-generated reference label rather than as an estimate of absolute labeling correctness. Collecting multiple independent annotations per tree and measuring inter-expert agreement would provide a stronger benchmark for future studies.

\paragraph{Sensitivity to fidelity of expert rule.}
To check whether the headline result is an artifact of how the
\texttt{NoDiff} predicate is phrased, we re-ran the full pipeline under
three additional variants of the same predicate, varying the inequality
($<$ vs $\le$) and the normalizer ($H_{\mathrm{field}}$ vs
\texttt{survey\_height}). All four variants were specified prior
to evaluation; we did not search for the rule that maximized the performance of the test
. Table~\ref{tab:rule-sensitivity} reports per-variant Macro-F1
on the same SRER+WREF test set; we report each variant individually rather
than summary statistics, since with $n{=}4$ a mean or percentile would
overstate what we can conclude. 

\begin{table}[!h]
  \caption{TreeAgent Macro-F1 under four rule phrasings of the
  \texttt{NoDiff} predicate, on the SRER+WREF test set. The headline rule
  is frozen \emph{a priori} for procedural fidelity to the annotation
  procedure.}
  \label{tab:rule-sensitivity}
  \begin{center}
    \begin{small}
      \begin{sc}
      \resizebox{\columnwidth}{!}{%
        \begin{tabular}{lc}
          \toprule
          Rule for \texttt{NoDiff}                                                                         & Macro-F1 \\
          \midrule
          $(\{H_{\mathrm{chm}}, H_{\mathrm{pcd}}\} - H_{\mathrm{field}})\,/\,H_{\mathrm{field}} < 0.02$    & 29.2\% \\
          $(\{H_{\mathrm{chm}}, H_{\mathrm{pcd}}\} - H_{\mathrm{field}})\,/\,H_{\mathrm{field}} \le 0.02$  & 21.9\% \\
          $(\{H_{\mathrm{chm}}, H_{\mathrm{pcd}}\} - H_{\mathrm{field}})\,/\,\texttt{survey\_height} < 0.02$ & 35.2\% \\
          $(\{H_{\mathrm{chm}}, H_{\mathrm{pcd}}\} - H_{\mathrm{field}})\,/\,\texttt{survey\_height} \le 0.02$ \textbf{(headline)} & \textbf{67.6\%} \\
          \midrule
          LightGBM baseline (Sec.~\ref{sec:ml-headline})                                                   & 36.2\% \\
          \bottomrule
        \end{tabular}%
        }
      \end{sc}
    \end{small}
  \end{center}
\end{table}

The four variants span 21.9--67.6\% Macro-F1, with
the headline rule, the last of the four. Given a faithfully-encoded rule, the headline rule clearly
outperforms the LightGBM baseline; the other three variants reach
21.9--35.2\% and either match or fall below 36.2\%. The four rules are mathematically distinct decision procedures: the headline rule remains the closest to the annotation procedure used to generate the reference labels, while the other three answer subtly different questions. The system's accuracy therefore depends on the fidelity between the encoded rule and the rule used by annotators. This is a deliberate design choice — D3 is built to faithfully execute expert rules, not to be robust to misspecification of them — and it places the burden on rule elicitation rather than on the executor.

\paragraph{Limitations of deterministic rules on edge cases.}
Despite operating as a deterministic rule-based system at
$\mathcal{V}_{\text{det}}$ nodes, TreeAgent misclassifies 2 trees in the
\texttt{NoDiff} class (Bias~6) under the headline rule. Both cases are
edge-case violations of the 2\% tolerance. Routing boundary instances to
a VLM node for adjudication is one option, but introducing VLM calls at
deterministic nodes would substantially increase per-tree cost and
latency. This points to an inherent tension in rule-driven agent
systems: hard symbolic thresholds are brittle at their boundaries, but
softening them with learned or visual adjudication carries a non-trivial
computational cost. The rarity of these cases suggests the current
design is a reasonable operating point; future work could consider
lightweight confidence scoring on deterministic predicates to flag
boundary cases for targeted VLM escalation.

\subsection{VLM Evaluation}
\label{sec:vlm-eval}

Table~\ref{tab:vlm-ablation} reports the \texttt{CrownOverlap} prompt version
results. All four versions tend to predict overlap: the \textbf{Minimal} prompt
achieves perfect recall (100\%) but only 7.1\% specificity, flagging 26 of 28
negative trees as positive. More detailed prompts reduce false positives: V1
cuts false positives from 26 to 21 and raises specificity to 25.0\%, while V3
reduces false positives to 17 and achieves 39.3\% specificity, the highest of
any version.

This comes at a recall cost. V3 misses 2 true overlaps that every other version
catches; V2 misses 1. We adopt \textbf{V1} as the default prompt for the
\texttt{CrownOverlap} node because it keeps recall high (100\%, zero missed
overlaps) while cutting false positives, and missing a real overlap is a worse
outcome than a false alarm when a human is reviewing the flags anyway. V3 yields the highest overall accuracy (62\%) but its missed cases
  mean real overlaps go unflagged.

When the model prediction clearly contradicts the image, this reflects a VLM limitation. General-purpose VLMs are pretrained predominantly on natural imagery and have not been exposed to LiDAR-specific rendering conventions~\citep{weng2025vlmremotesensing}, making it harder to interpret complicated crown boundaries where the decision to label crown overlap is marginal. Fine-tuning an image model on domain-specific LiDAR data is a natural next step. We also note that some ground-truth labels in this set may contain human labeling error.

\begin{table}[h!]
  \caption{%
    Prompt version results on $n{=}50$ OSBS trees (22 positive, 28 negative).
    Model: \texttt{claude-sonnet-4-6}.
    FP\,=\,false positives; FN\,=\,false negatives.
  }
  \label{tab:vlm-ablation}
  \begin{center}
    \begin{small}
      \begin{tabular}{lcccc}
        \toprule
        Version & Accuracy & Recall  & Specificity & FP / FN \\
        \midrule
        Minimal & 48.0\%   & 100.0\% &  7.1\%      & 26 / 0  \\
        V1      & 58.0\%   & 100.0\% & 25.0\%      & 21 / 0  \\
        V2      & 52.0\%   &  95.5\% & 17.9\%      & 23 / 1  \\
        V3      & 62.0\%   &  90.9\% & 39.3\%      & 17 / 2  \\
        \bottomrule
      \end{tabular}
    \end{small}
  \end{center}
\end{table}


\subsection{ML Baseline: Headline 7-Class Result}
\label{sec:ml-headline}

On the geographic split (train OSBS $n=136$, test WREF $n=79$ + SRER $n=68$), LightGBM with class-weighted training reaches a test \textbf{macro-F1 of 36.2\%} on the seven-class bias taxonomy. Per-site, the model attains 40.8\% on WREF and 14.2\% on SRER (Table~\ref{tab:ml-headline}). Train macro-F1 reaches 86\%: the model fits OSBS well, but the fit does not fully transfer to either test ecosystem. The dominant per-class failure is \texttt{bias\_2} ($H_{\mathrm{field}}$ overestimation) being predicted as \texttt{bias\_4} ($H_{\mathrm{chm}}$ underestimation only). Both classes produce $H_{\mathrm{chm}} < H_{\mathrm{field}}$ and are indistinguishable from summary statistics, discriminating between them requires spatial evidence (was a neighboring crown inflating the field reading? did rasterization miss the apex?) that the tabular features discard. The \texttt{unknown} class is also hard to predict: it captures trees the experts could not place in any mechanism category within five minutes, so the class spans the same feature space as the other six rather than occupying a coherent region.

\begin{table}[!h]
  \caption{Supervised ML baseline on the seven-class bias taxonomy. Train: OSBS ($n=136$). Test: WREF ($n=79$), SRER ($n=68$). Numbers are test macro-F1.}
  \label{tab:ml-headline}
  \begin{center}
    \begin{small}
      \begin{sc}
        \begin{tabular}{lc}
          \toprule
          Site                              & Test macro-F1 $\uparrow$ \\
          \midrule
          WREF                              & 40.8\% \\
          SRER                              & 14.2\% \\
          \textbf{Combined (WREF + SRER)}   & \textbf{36.2\%} \\
          \bottomrule
        \end{tabular}
      \end{sc}
    \end{small}
  \end{center}
  \vskip -0.1in
\end{table}

\paragraph{Robustness of the headline.} The 36.2\% macro-F1 result is stable
across the configuration space; no single intervention pushes performance
meaningfully past it. Five interaction features targeting the dominant
\texttt{bias\_2}/\texttt{bias\_4} confusion lift macro-F1 to 37.3\%;
SMOTE~\citep{chawla2002smote} and its Borderline and ADASYN variants drop it
to 30.0\%--34.1\%; focal-style reweighting~\citep{lin2017focal} reaches
34.5\%; transductive and \texttt{unknown}-only
pseudo-labeling~\citep{lee2013pseudo} stay in 34.9\%--35.1\%. Model-family
substitutions (CatBoost 35.3\%; HistGradientBoosting 26.3\%;
LightGBM$+$RF$+$XGBoost stacking~\citep{breiman2001random,chen2016xgboost}
20.9\%) and a tabular foundation model
(TabPFN~\citep{hollmann2025tabpfn}, 24.6\%) all underperform LightGBM. Adding
the per-tree point-cloud maximum $H_{\mathrm{pcd}}$ together with seven
derived percentile statistics drops macro-F1 to 32.4\%, and
PointNet~\citep{qi2017pointnet} on the cropped raw point clouds underperforms
the expert-feature baseline. The largest single performance lift comes from
abandoning the cross-site split: a pooled classifier on stratified 20\%
per-site holdouts reaches $45.1\% \pm 7.0\%$ macro-F1 over five seeds, an
$+8.9$-point lift over the 36.2\% headline. Cross-site ecosystem shift
therefore accounts for roughly a quarter of the gap to within-site
performance, with the remainder reflecting absolute training-set size.
Per-configuration numbers, the reversed-split robustness check, and full
ablation are in Appendix~\ref{app:ml-ablations}.

\paragraph{Limitations.} Four caveats apply. First, $n_{\mathrm{train}} = 136$
trees come from a single site (OSBS); the numbers reflect a worst-case
single-site training scenario. Second, the training set is too small for
reliable held-out hyperparameter tuning, so we fix hyperparameters by hand.
Third, results depend on the specific site pairing of the geographic split;
the reversed-split numbers in Appendix~\ref{app:ml-ablations} are the closest
available robustness check. Fourth, we report point estimates without
confidence intervals; bootstrap CIs would tighten the per-site comparisons
given the single-digit test counts on some bias categories.

\section{Conclusion}

We presented TreeAgent, a multi-agent framework for automated tree-level bias labeling in forest remote sensing that combines expert decision rules with Vision-Language Models under the Decoupled Declarative Decision (D3) framework. TreeAgent achieves substantially higher agreement with expert-provided reference labels than the supervised ML baseline (67.6\% vs.\ 36.2\% Macro-F1) and reproduces expert-defined labels at 0.040 minutes per tree, compared with approximately 5 minutes of expert annotation effort. These results demonstrate that agentic orchestration of structured expert priors with VLMs is a viable path toward scalable, interpretable annotation in domains where ground truth is expensive.

Two complementary findings characterize the system's current ceiling. First, \textbf{VLM output is the dominant source of labeling error}: recall is highest in classes resolved by deterministic nodes alone and degrades progressively with VLM involvement. This bottleneck is not a structural flaw of D3, the deterministic layer performs reliably, but a reflection of the limits of current general-purpose VLMs on specialized forestry imagery. Second, the deterministic layer itself exhibits \textbf{brittleness at threshold boundaries}: two no-difference trees that nominally fall under a pure arithmetic rule are misclassified because their measurements sit at the edge of the 2\% tolerance, a regime where no symbolic rule can yet substitute for genuine perceptual judgment.

These findings suggest two directions for future work. On the VLM side, domain adaptation, through fine-tuning on labeled CHM and LiDAR transect imagery or retrieval-augmented prompting with similar resolved cases, could close the accuracy gap in visually ambiguous classes without changing the D3 framework.
On the rule side, augmenting the \texttt{NoDiff} and other threshold predicates
with lightweight confidence scores could enable selective VLM escalation,
targeting only the cases where the deterministic predicate is genuinely
uncertain. While evaluated only in forestry, D3 suggests a possible recipe for scientific labeling workflows in domains where expert reasoning is structured and evolving but
occasional perceptual judgment is unavoidable. By separating what to decide from how to decide it, the framework accommodates rule revisions without
re-engineering and produces decisions that are auditable against the same
diagnostic logic domain experts apply---properties that matter as much for
trust and reproducibility as for accuracy in high-stakes annotation pipelines.

\section*{Impact Statement}

This work aims to scale expert-driven scientific labeling in forest remote 
sensing, where tree-height bias correction underpins biomass and carbon 
estimates used in climate policy. By orchestrating expert rules with VLMs, 
TreeAgent reduces annotation cost while preserving auditability---each 
decision traces to an expert rule rather than opaque model weights. To our 
knowledge, this is the first agent system designed for this purpose in 
forestry, with the potential to quantify height-induced biases in tree 
biomass estimation at national scale far faster than current expert 
workflows allow. Although demonstrated only in forestry, the framework may be applicable to other domains where expert decision procedures can be formalized into structured rules. We caution 
that the system's labels are not a substitute for expert review in 
high-stakes inventories: VLM errors on perceptual nodes propagate to 
downstream carbon estimates, and the D3 framework inherits whatever biases 
exist in the encoded expert rules.

\bibliography{reference}
\bibliographystyle{icml2026}

\appendix
\onecolumn
\section{D3 Framework Neural Rule Transpiler Prompt}
\label{app:d3}
\textit{System prompt:}
\begin{lstlisting}
You are a converter that translates an expert-defined classification rule (written in markdown) into a JSON TreeConfig for a multi-agent tree classifier.
 
Available Node Classes
 
Each node has a fixed id, type, fields, and condition. You must use ONLY these classes:
 
Deterministic nodes (evaluate numeric conditions from CSV data and/or presence of data)
- HeightComparatorChmPcd     -> id: "chm_pcd"              | true if h_chm < h_pcd
# ...
# Full node-class definitions withheld pending separate publication.
 
VLM nodes (require visual inspection of images)
- CrownOverlap               -> id: "crown_overlap"         | true if crown overlap detected
- GroundPcdOutlier           -> id: "ground_pcd_outlier"    | true if ground point cloud outlier detected
 
End nodes (classification result)
- FieldUnderestimation       -> id: "1"   | label: H_field Underestimation
- FieldOverestimation        -> id: "2"   | label: H_field Overestimation
- ChmOverestimation      -> id: "3"   | label: H_chm Overestimation
- ChmUnderestimation          -> id: "4"   | label: H_chm Underestimation
- ChmPcdUnderestimation         -> id: "5"   | label: H_chm and H_pcd Underestimation
- NoDifference                     -> id: "6"   | label: No Difference
- Unknown                    -> id: "7"   | label: Unknown
 
Output Format
 
Output ONLY valid JSON, **no explanation, no markdown fences**. The format is:
 
{
  "tree_id": "<tree_id provided by user>",
  "root": "<node_id of the first node>",
  "nodes": {
    "<node_id>": {
      "class": "<NodeClassName>",
      "edges": {
        "True": "<next_node_id>",
        "False": "<next_node_id>"
      }
    },
    ...
    "<end_node_id>": {
      "class": "<EndNodeClassName>"
    }
  }
}
 
Rules
- End nodes do NOT have edges.
- Every non-end node MUST have edges with "true" and "false" keys.
- All node ids referenced in edges must exist as keys in "nodes".
- Use the exact node ids listed above (e.g. "chm_pcd", "crown_overlap", "1", "7").
- End nodes (type "end") may share ids across the tree since they have no edges.
- Non-end nodes must have globally unique ids. If the same logical check appears 
  at multiple positions with different children, append a suffix to distinguish them,
  e.g. "chm_pcd_1", "chm_pcd_2". The "class" field still indicates which Node class to use.
- tree_id must be exactly the string passed in by the user.
\end{lstlisting}
\textit{User prompt:}
\begin{lstlisting}
tree_id: {tree_id}

Expert rule:
        {rule_markdown}

Output ONLY valid JSON, no explanation, no markdown fences.
\end{lstlisting}

\section{CrownOverlap Prompt Variants}
\label{app:prompts}

\subsection*{Minimal}

\textit{System prompt:} none.

\textit{User turn:}
\begin{lstlisting}
Is there crown overlap between the target tree and any neighboring tree?

Return ONLY this JSON:
{
  "crown_overlap_detected": <true|false>,
  "confidence": <0.0-1.0>
}
\end{lstlisting}

\subsection*{V1}

\textit{System prompt:}
\begin{lstlisting}
You are analyzing two LiDAR figures of a forest plot.

Figure 1 - bird's-eye canopy height model (CHM):
  - Background color shows canopy height. Warmer / lighter = taller canopy.
  - Dashed circles show individual tree crown boundaries.
  - Brown dots show stem locations (trunk base).
  - The target tree is identified by its ID in the image title.

Figure 2 - transect cross-section (two side-by-side panels):
  - Left panel: N-S cross-section. Right panel: E-W cross-section.
  - Point cloud colored by LiDAR class:
      Dark gray = ground returns.
      Yellow / orange = low vegetation.
      Green = high vegetation / canopy.
  - Red circles mark survey measurement points.

Output ONLY valid JSON. No prose, no markdown fences.
\end{lstlisting}

\textit{User turn:}
\begin{lstlisting}
Does any neighboring tree's crown circle overlap the target tree's crown circle?

Return exactly this JSON:
{
  "crown_overlap_detected": <true|false>,
  "overlap_description": "<brief description of what you see>",
  "confidence": <0.0-1.0>
}
\end{lstlisting}

\subsection*{V2}

\textit{System prompt:} same as V1.

\textit{User turn:}
\begin{lstlisting}
Does any neighboring tree's crown circle intersect or overlap the target tree's
crown circle, and is that neighbor taller?

Step 1 - Read Figure 1 axes and color scale.
Step 2 - Locate the target tree's dashed crown circle in Figure 1.
Step 3 - Do any OTHER crown circles intersect or overlap the TARGET CROWN
         CIRCLE? (Circles cross or one partially covers the other. NOT
         whether a circle covers a stem dot.)
Step 4 - If overlap found: is that neighbor's CHM color warmer / lighter?
Step 5 - In Figure 2: are there green (Class 5) returns at a HIGHER elevation
         than the target apex, within the lateral extent of the target crown?

Return exactly this JSON:
{
  "crown_overlap_detected": <true|false>,
  "taller_neighbor_detected": <true|false>,
  "overlap_description": "<2-3 sentences citing specific visual evidence>",
  "confidence": <0.0-1.0>
}

Rules:
  - crown_overlap_detected = true if dashed circles cross, regardless of
    stem positions.
  - taller_neighbor_detected = true only if the overlapping neighbor is
    visually taller.
  - When evidence is ambiguous, default both to false and set confidence < 0.5.
\end{lstlisting}

\subsection*{V3}

\textit{System prompt:}
\begin{lstlisting}
You are an experienced forestry and LiDAR expert. Your task is to determine
whether a target tree's height measurement could be inflated by a taller
neighboring tree's crown.

Figure 1 - bird's-eye canopy height model (CHM):
  - Background color shows canopy height. Warmer / lighter = taller canopy.
  - Dashed circles show individual tree crown boundaries.
  - Brown dots show stem locations (trunk base).
  - The target tree is identified by its ID in the image title.

Figure 2 - transect cross-section (two side-by-side panels):
  - Left panel: N-S transect (X = Northing, Y = elevation in meters).
  - Right panel: E-W transect (X = Easting, Y = elevation in meters).
  - Point cloud colored by LiDAR class:
      Class 1 (dark gray): ground returns.
      Class 2 (yellow / orange): low vegetation.
      Class 5 (green): high vegetation / canopy.
  - Red circles mark survey measurement points.

Output ONLY valid JSON. No prose, no markdown fences.
\end{lstlisting}

\textit{User turn:}
\begin{lstlisting}
Determine whether any neighboring tree's crown circle intersects or overlaps
the target tree's crown circle, and whether any such neighbor is taller.

Step 1 - Read Figure 1 axes and scale (Easting, Northing, height color range).
Step 2 - Locate the target tree's dashed crown circle; note if identification
         is clear or ambiguous.
Step 3 - From Figure 1: do any OTHER crown circles intersect or overlap the
         TARGET CROWN CIRCLE? (Circles cross or one partially covers the
         other -- NOT whether a circle covers a stem dot.) If yes, is that
         neighbor's CHM color warmer / lighter (taller)?
Step 4 - From Figure 2: locate the grey cylinder (target trunk/crown). Are
         there substantial green (Class 5) returns at a HIGHER elevation than
         the target apex within the lateral extent of the target crown?

Return exactly this JSON:
{
  "crown_overlap_detected": <true|false>,
  "taller_neighbor_detected": <true|false>,
  "overlap_description": "<2-3 sentences citing specific visual evidence>",
  "confidence": <0.0-1.0>
}

Rules:
  - crown_overlap_detected = true if dashed circles cross or one partially
    covers the other, regardless of stem positions.
  - taller_neighbor_detected = true only if such a neighbor is visually taller.
  - When evidence is ambiguous, default both to false and set confidence < 0.5.
\end{lstlisting}

\section{Full feature list for the supervised ML baseline}
\label{app:features}

Table~\ref{tab:full-features} lists all 25 features used by the supervised ML baseline (Section~\ref{sec:ml-baseline}), with their source and a brief description.

\begin{table}[h]
\caption{Complete feature list for the supervised ML baseline. \emph{Source} indicates whether the feature comes from the NEON Vegetation Structure field survey (Survey), the canopy height model raster (CHM), the LiDAR point cloud (PCD), or is derived from combinations of the above (Derived).}
\label{tab:full-features}
\centering
\small
\begin{tabular}{llp{0.55\textwidth}}
\toprule
Name & Source & Description \\
\midrule
\texttt{height}                       & Survey  & Field-tape tree height $H_{\mathrm{field}}$ (m) \\
\texttt{stemDiameter}                 & Survey  & Diameter at breast height (cm) \\
\texttt{crownRadius}                  & Survey  & Crown radius (m) \\
\texttt{adjElevation}                 & Survey  & Adjusted ground elevation reported in the field record (m) \\
\texttt{canopyPosition}               & Survey  & Categorical: dominant, codominant, intermediate, suppressed, open \\
\texttt{plantStatus}                  & Survey  & Categorical: live, dead, broken, etc. \\
\texttt{growthForm}                   & Survey  & Categorical: single bole tree, multi-bole tree, sapling, etc. \\
\texttt{taxonID}                      & Survey  & Categorical: NEON species code (e.g.\ PIPA2, QULA2) \\
\addlinespace
\texttt{chm\_height}                  & CHM     & Single-pixel CHM value at the field-survey coordinate (m) \\
\texttt{chm\_height\_buff}            & CHM     & Maximum CHM value within a buffer of radius $0.9 \times \texttt{crownRadius}$ around the stem (m) \\
\addlinespace
\texttt{Z\_pointcloud}                & PCD     & Mean elevation of LiDAR returns classified as ground (LAS class 2) inside the per-tree buffer (m) \\
\texttt{std\_pointcloud}              & PCD     & Standard deviation of those ground-point elevations (m) \\
\texttt{pointdensity\_ground\_tree}   & PCD     & Ground-point density inside the per-tree buffer (points/m$^2$) \\
\texttt{dtm\_nn}                      & PCD     & Elevation reported by the nearest defined Digital Terrain Model grid cell (m) \\
\texttt{dtm\_nn\_dist}                & PCD     & Distance from the field-survey coordinate to that nearest DTM cell (m); large values flag unreliable terrain interpolation \\
\texttt{dtm\_buffer}                  & PCD     & Mean DTM elevation inside the per-tree buffer (m) \\
\addlinespace
\texttt{diff\_chm\_survey}            & Derived & $\texttt{chm\_height} - \texttt{height}$ (m) \\
\texttt{diff\_chm\_survey\_pct}       & Derived & $\texttt{diff\_chm\_survey} / \texttt{height}$ (unitless) \\
\texttt{abs\_diff\_chm\_survey}       & Derived & $|\texttt{diff\_chm\_survey}|$ (m) \\
\texttt{diff\_chm\_buff\_survey}      & Derived & $\texttt{chm\_height\_buff} - \texttt{height}$ (m) \\
\texttt{ground\_elevation\_diff}      & Derived & $\texttt{adjElevation} - \texttt{Z\_pointcloud}$ (m) \\
\texttt{ground\_std\_to\_density}     & Derived & $\texttt{std\_pointcloud} / \texttt{pointdensity\_ground\_tree}$ \\
\texttt{dtm\_nn\_vs\_buffer}          & Derived & $\texttt{dtm\_nn} - \texttt{dtm\_buffer}$ (m); large values mean DTM disagrees with local LiDAR \\
\texttt{height\_to\_crown\_ratio}     & Derived & $\texttt{height} / \texttt{crownRadius}$ (canopy slenderness) \\
\texttt{height\_to\_dbh\_ratio}       & Derived & $\texttt{height} / \texttt{stemDiameter}$ \\
\bottomrule
\end{tabular}
\end{table}

The five interaction features evaluated in the engineered-features ablation (Section~\ref{app:ml-ablations}) build on this base set: \texttt{chm\_apex\_drift} $= \texttt{diff\_chm\_survey} - \texttt{diff\_chm\_buff\_survey}$ (apex offset from the stem), \texttt{ground\_noise} $= \texttt{std\_pointcloud} / \texttt{pointdensity\_ground\_tree}$ (local terrain noise), \texttt{rel\_chm\_gap} $= \texttt{diff\_chm\_survey} / \texttt{height}$ (scale-invariant gap), \texttt{shape\_x\_gap} $= \texttt{height\_to\_crown\_ratio} \times |\texttt{diff\_chm\_survey}|$ (slenderness $\times$ gap), and \texttt{dtm\_dist\_x\_gap} $= \texttt{dtm\_nn\_dist} \times |\texttt{diff\_chm\_survey}|$ (DTM unreliability $\times$ gap).

\section{Tabular ML ablations: full results}
\label{app:ml-ablations}

This appendix contains the per-configuration ablation numbers summarized in
Section~\ref{sec:ml-headline}. All numbers are test macro-F1 on the
OSBS-train, WREF$+$SRER-test geographic split unless otherwise noted; per-site
columns give WREF and SRER.

\paragraph{Reversed split.} Swapping train and test (train WREF$+$SRER, test
OSBS) yields a 7-class macro-F1 of 29.1\%. Cross-site difficulty is roughly
symmetric in magnitude: the gap is not an artifact of training on the smallest
site, even though OSBS is slightly easier as a training set than as a test set.

\paragraph{Within-site evaluation.} The cross-site geographic split is the
deployment-relevant question, but it conflates two sources of difficulty:
small training sets and structural shift between ecosystems. To isolate the
structural-shift cost, we hold out a stratified 20\% test subset within each
of the three sites, train on the union of the within-site train sets, and
evaluate on the union of the test sets (one pooled classifier). Macro-F1
averaged over five seeds reaches \textbf{$45.1\% \pm 7\%$} --- a $+8.9\%$
lift over the cross-site headline of 36.2\%. The result is stable across
split fractions $\{10\%, 20\%, 30\%, 40\%\}$ (within $\pm 0.5\%$ of the 20\%
number).

\paragraph{Site-specific classifiers.} Training one LightGBM per site (using
the same within-site 80/20 split) and reporting the macro-average of the
three per-site test scores reaches only $34.6\% \pm 4.5\%$. Pooling across
sites lifts macro-F1 by 10\% over per-site fitting, even when test data is
held out within-site. The signal is therefore in shared structure across
ecosystems rather than in site-specific patterns the per-site classifiers
could learn from $\sim 65$--$109$ training trees apiece.

\paragraph{Engineered features.} Five interaction features target the
\texttt{bias\_2} vs.\ \texttt{bias\_4} confusion that dominates the per-class
error: \texttt{chm\_apex\_drift} (the difference between the buffered-crown
CHM maximum and the at-stem CHM value, flagging trees whose tallest pixel is
offset from the survey coordinate), \texttt{ground\_noise} (the local LiDAR
ground-elevation standard deviation divided by ground-point density, flagging
noisy DTM regions), \texttt{rel\_chm\_gap} ($\texttt{diff\_chm\_survey} /
H_{\mathrm{field}}$, scale-invariant across short and tall trees),
\texttt{shape\_x\_gap} (\texttt{height\_to\_crown\_ratio} multiplied by the
absolute CHM--field gap, coupling canopy slenderness with disagreement
magnitude), and \texttt{dtm\_dist\_x\_gap} (\texttt{dtm\_nn\_dist} multiplied
by the absolute CHM--field gap, up-weighting gaps where the DTM is
unreliable). Adding these to the headline feature set raises 7-class macro-F1
from 36.2\% to 37.3\% and lifts SRER macro-F1 from 14.2\% to 16.7\%. The lift is
small because the features target the \texttt{bias\_2}/\texttt{bias\_4}
confusion and do not help the \texttt{unknown} class.

\paragraph{Class imbalance.} The 4-to-46 imbalance across the seven classes
invites two strategy families: oversampling and loss reweighting.
SMOTE~\citep{chawla2002smote} interpolates new minority-class points between
real ones; we test it together with its borderline and ADASYN variants. All
three drop 7-class macro-F1 (30.4\%--34.1\%), because the dominant failure is
class overlap rather than imbalance: synthetic \texttt{bias\_5} samples
interpolated between two real \texttt{bias\_5} trees still sit inside the
\texttt{bias\_4} region. Focal-loss-style reweighting~\citep{lin2017focal}
reweights each sample's contribution to the loss by $(1/p_c)^{1.5}$, where
$p_c$ is the class frequency; this also drops macro-F1 (0.345). Inverse-
frequency class weighting (already used in the headline) handles imbalance
well enough on this data; aggressive minority oversampling does more harm
than good.

\paragraph{Cross-site adaptation.} Transductive
pseudo-labeling~\citep{lee2013pseudo} adds the model's most confident test
predictions back into training and refits. On the 7-class target this
slightly degrades macro-F1 (34.9\%) but raises SRER macro-F1 (17.1\%). A
leak-free variant pseudo-labels only the 23 training-set rows tagged
\texttt{unknown} (using a model fit on the labeled rows only), holding all
test data out, and reaches macro-F1 35.1\% with WREF macro-F1 43.0\%. Both
variants help WREF and SRER unevenly without lifting the overall 7-class
number, since the headline classifier's confident predictions on
\texttt{unknown} test rows propagate into the augmented training set as
noise.

\paragraph{Model family.} CatBoost (35.3\%) and HistGradientBoosting (26.3\%)
underperform LightGBM. A stacking ensemble (LightGBM, RF, and XGBoost feeding
a logistic regression meta-learner with out-of-fold predictions) underperforms
further (20.9\%), because the meta-learner overfits small out-of-fold
predictions on classes with as few as four training examples.

\paragraph{Tabular foundation model.} TabPFN~\citep{hollmann2025tabpfn} run
end-to-end reaches macro-F1 24.6\%, well below LightGBM. Its
synthetic-tabular prior does not capture the cross-ecosystem structural
shift between OSBS, WREF, and SRER. Using TabPFN to pseudo-label the most
confident 30\% of test rows and retraining LightGBM on the union (21.6\%)
does not recover; on the 7-class target the prior's confident wrong
predictions on the \texttt{unknown} class propagate as label noise.

\paragraph{Adding raw point-cloud canopy maximum.} Extracting the per-tree
maximum canopy height ($H_{\mathrm{pcd}}$) and adding it together with seven
derived point-cloud statistics (98th and 95th percentile heights, mean and
standard deviation of return heights, canopy and ground point counts, and
the mean ground elevation) drops 7-class macro-F1 to 32.4\%. Two factors
explain this: partial redundancy with the ground-quality features the
expert labeling process already encodes, plus overfitting from adding eight
new features to a 136-row training set. The result reinforces D3's framing:
even with the canopy maximum in hand, summary statistics do not resolve the
mechanism distinction between \texttt{bias\_2} ($H_{\mathrm{field}}$
overestimation), \texttt{bias\_4} ($H_{\mathrm{chm}}$ raster
underestimation), and \texttt{bias\_5} ($H_{\mathrm{chm}}$ and
$H_{\mathrm{pcd}}$ both underestimate).

\paragraph{Raw-data baseline (PointNet).}
PointNet~\citep{qi2017pointnet} trained directly on the cropped point clouds
underperforms the expert-feature LightGBM on the 7-class target. Every grid
configuration peaks within the first 10--50 epochs and then overfits, and
larger hidden widths underperform smaller ones. Concatenating PointNet's
max-pooled features with the tabular features and feeding them to LightGBM
\emph{degrades} the tabular-only baseline. At $n_{\mathrm{train}}\sim 100$,
the engineered features encode ecological domain knowledge the network does
not recover from labels at this scale.

\begin{table}[t]
  \caption{Full ablation summary on the seven-class bias taxonomy. All numbers are test macro-F1 on the geographic split (OSBS train, WREF$+$SRER test) unless noted; per-site columns give WREF and SRER. The headline number is the supervised baseline (top row).}
  \label{tab:ml-ablations}
  \begin{center}
    \begin{small}
      \begin{sc}
\begin{tabular}{lccc}
          \toprule
          Configuration                                       & macro-F1 $\uparrow$ & WREF & SRER \\
          \midrule
          \textbf{7-class baseline (headline)}                & \textbf{36.2\%} & 40.8\% & 14.2\% \\
          \quad + engineered features                         & 37.3\% & 40.5\% & 16.7\% \\
          \quad + $H_{\mathrm{pcd}}$ point-cloud features     & 32.4\% & 33.3\% & 13.6\% \\
          \midrule
          \multicolumn{4}{l}{\emph{Reversed split (train WREF$+$SRER, test OSBS only)}} \\
          \quad 7-class                                       & 29.1\% & --- & --- \\
          \midrule
          \multicolumn{4}{l}{\emph{Within-site split (mean $\pm$ std over 5 seeds, 80/20 test)}} \\
          \quad Pooled (one classifier, all three sites)      & $45.1\% \pm 7.0\%$ & 47.3\% & 30.5\% \\
          \quad Site-specific (per-site, macro-avg)           & $34.6\% \pm 4.5\%$ & 40.0\% & 32.1\% \\
          \midrule
          \multicolumn{4}{l}{\emph{Class imbalance on the original split}} \\
          \quad + focal-style $(1/p_c)^{1.5}$ reweighting     & 34.5\% & 39.3\% & 12.0\% \\
          \quad + SMOTE                                       & 31.5\% & 39.3\% & 8.7\% \\
          \quad + Borderline-SMOTE                            & 30.4\% & 34.8\% & 11.0\% \\
          \quad + ADASYN                                      & 34.1\% & 44.1\% & 7.9\% \\
          \midrule
          \multicolumn{4}{l}{\emph{Cross-site adaptation}} \\
          \quad Pseudo-labeling on test (transductive)        & 34.9\% & 37.8\% & 17.1\% \\
          \quad Pseudo-labeling on \texttt{unknown} pool      & 35.1\% & 43.0\% & 11.3\% \\
          \midrule
          \multicolumn{4}{l}{\emph{Model family}} \\
          \quad CatBoost                                      & 35.3\% & 39.3\% & 14.3\% \\
          \quad HistGradientBoosting                          & 26.3\% & 30.3\% & 11.6\% \\
          \quad Stacking (LGBM+RF+XGB$\to$LR)                 & 20.9\% & 28.2\% & 11.8\% \\
          \quad TabPFN end-to-end                             & 24.6\% & 30.2\% & 13.2\% \\
          \quad TabPFN pseudo $\to$ LightGBM                  & 21.6\% & 23.5\% & 13.5\% \\
          \bottomrule
        \end{tabular}
      \end{sc}
    \end{small}
  \end{center}
\end{table}

\end{document}